\documentclass[conference]{IEEEtran}
\IEEEoverridecommandlockouts

\usepackage{cite}
\usepackage{amsmath, amssymb, amsfonts}
\usepackage{algorithmic}
\usepackage{graphicx}
\usepackage{textcomp}
\usepackage{xcolor}
\usepackage[compact]{titlesec}
\def\BibTeX{{\rm B\kern-.05em{\sc i\kern-.025em b}\kern-.08em T\kern-.1667em\lower.7ex\hbox{E}\kern-.125emX}}
\usepackage{url}
\usepackage{subcaption}
\usepackage{hyperref}

\usepackage{tabularx}
\IEEEaftertitletext{\vspace{-2\baselineskip}}
\titlespacing*{\section}{0pt}{*0.5}{*0.5} 
\titlespacing*{\subsection}{0pt}{*0.5}{*0.4}
\usepackage[font=footnotesize, labelfont=bf]{caption}

\IEEEoverridecommandlockouts
\IEEEpubid{\makebox[\columnwidth]{979-8-3315-8654-6/25/\$31.00~\copyright2025 IEEE \hfill} \hspace{\columnsep}\makebox[\columnwidth]{ }}

\begin{document}
    \title{Genetic Algorithms For Parameter Optimization for Disparity Map
    Generation of Radiata Pine Branch Images\\
    }
    \author{\IEEEauthorblockN{Yida Lin, Bing Xue, Mengjie Zhang} \IEEEauthorblockA{\small \textit{Centre for Data Science and Artificial Intelligence} \\ \textit{Victoria University of Wellington, Wellington, New Zealand}\\ linyida\texttt{@}myvuw.ac.nz, bing.xue\texttt{@}vuw.ac.nz, mengjie.zhang\texttt{@}vuw.ac.nz}
    \and \IEEEauthorblockN{Sam Schofield, Richard Green} \IEEEauthorblockA{\small \textit{Department of Computer Science and Software Engineering} \\ \textit{University of Canterbury, Canterbury, New Zealand}\\ sam.schofield\texttt{@}canterbury.ac.nz, richard.green\texttt{@}canterbury.ac.nz}
    }

    \maketitle
    \IEEEpubidadjcol
    \begin{abstract}
        Traditional stereo matching algorithms like Semi-Global Block Matching (SGBM)
        with Weighted Least Squares (WLS) filtering offer speed advantages over neural
        networks for UAV applications, generating disparity maps in
        approximately 0.5~seconds per frame. However, these algorithms require
        meticulous parameter tuning. We propose a Genetic Algorithm (GA) based parameter
        optimization framework that systematically searches for optimal parameter
        configurations for SGBM and WLS, enabling UAVs to measure distances to
        tree branches with enhanced precision while maintaining processing
        efficiency. Our contributions include: (1) a novel GA-based parameter
        optimization framework that eliminates manual tuning; (2) a comprehensive
        evaluation methodology using multiple image quality metrics; and (3) a
        practical solution for resource-constrained UAV systems. Experimental results
        demonstrate that our GA-optimized approach reduces Mean Squared Error by
        42.86\% while increasing Peak Signal-to-Noise Ratio and Structural
        Similarity by 8.47\% and 28.52\%, respectively, compared with baseline
        configurations. Furthermore, our approach demonstrates superior
        generalization performance across varied imaging conditions, which is critcal
        for real-world forestry applications.
    \end{abstract}


    \section{Introduction}
    The timber industry represents a significant component of New Zealand's
    economy, with Radiata Pine being a predominant commercial species across the
    country's plantation forests\cite{lin2024deep}. Regular pruning of Radiata Pine
    is essential for enhancing wood quality, promoting straight trunk growth, and
    accelerating timber production. Traditionally, this pruning process requires
    workers to physically climb trees, which presents substantial safety hazards
    due to the considerable heights involved and, in some locations, proximity
    to high-voltage power lines. To address these safety concerns and increase operational
    efficiency, Unmanned Aerial Vehicles (UAVs) have emerged as promising tools
    for automating forestry maintenance tasks, particularly branch pruning \cite{lin2024drone}.
    UAVs eliminate the need for human climbers, thereby reducing workplace accidents
    while maintaining or improving pruning precision. Our research team has
    developed an autonomous drone system equipped with stereo cameras for branch
    detection and a specialized pruning mechanism capable of accurately trimming
    branches at various heights. This approach not only enhances worker safety but
    also increases pruning efficiency in commercial Radiata Pine plantations.

    Accurate distance estimation to tree branches from stereo camera-generated depth
    maps is crucial for successful UAV-assisted pruning operations. Previous
    research has explored integrating Semi-Global Block Matching (SGBM)
    \cite{hirschmuller2007stereo} with segmentation models for branch detection
    and distance measurement in radiata pine forests \cite{lin2024drone}. Furthermore,
    combining stereo vision with You Only Look Once (YOLO)-based object
    detection \cite{lin2024deep} has shown promising results in precise branch
    localization, enhancing both identification and pruning efficiency.

    While neural network-based approaches \cite{laga2020survey}, such as Pyramid
    Stereo Matching Network (PSMNet) \cite{chang2018pyramid}, provide high-precision
    depth estimation, their significant computational demands result in low
    frame rates (approximately 1~second per frame), making them impractical for real-time
    UAV applications to radiata pine branch pruning. In contrast, the traditional
    SGBM method \cite{gehrig2009real} generates depth maps much faster (approximately
    0.5~seconds per frame) but requires meticulous parameter tuning to balance speed
    and accuracy. SGBM also offers superior interpretability compared with
    neural networks through its transparent pixel-wise matching and structured
    cost aggregation processes—a critical advantage for safety-critical UAV
    operations where understanding algorithm behavior is essential. This structured
    methodology makes SGBM particularly suitable for UAV systems with limited computational
    resources.

    Despite SGBM's advantages, its performance heavily depends on parameters
    traditionally tuned manually based on experience. To overcome this limitation,
    we propose a Genetic Algorithm (GA) \cite{coello2000updated} framework to
    optimize SGBM and Weighted Least Squares (WLS) \cite{farbman2008edge} that
    systematically adjusts key parameters to enhance depth map accuracy while
    maintaining computational efficiency. GA is particularly well-suited for this
    task because it excels in global search for complex, high-dimensional
    optimization, efficiently exploring a large parameter space and avoiding local
    minima \cite{coello2000updated}. By iteratively refining candidate solutions,
    our GA approach ensures systematic parameter optimization, achieving swift
    convergence while balancing accuracy and computational cost.

    The \textbf{primary objectives} of this research are to: (1) develop an
    automated parameter optimization framework for SGBM+WLS that eliminates the
    need for manual parameter tuning; (2) improve the accuracy of disparity maps
    for tree branch detection while maintaining processing efficiency at 0.5 seconds
    per frame; and (3) verify the effectiveness of GA in optimizing stereo
    matching algorithms.

    Our experimental results demonstrate that the proposed GA-optimized SGBM and
    WLS framework significantly reduces the error between generated disparity
    maps and gold standard. This is evidenced by enhanced evaluation metrics: a
    42.86\% lower Mean Squared Error (MSE) \cite{sara2019image} indicates reduced
    differences between estimated and true depth values, while higher Peak
    Signal-to-Noise Ratio (PSNR) \cite{sara2019image} and SSIM values (increases
    of 8.47\% and 28.52\%, respectively) reflect improved structural consistency
    and more effective perceptual detail capture \cite{sara2019image}. Moreover,
    our method maintains high accuracy under varying lighting conditions and complex
    backgrounds, underscoring its robustness for UAV applications.

    The key contributions of this work include:
    \begin{itemize}
        \item A novel GA-based optimization framework for automatically tuning SGBM
            and WLS parameters that eliminates the need for manual configuration;

        \item A comprehensive evaluation methodology using multiple image quality
            metrics (MSE, PSNR, SSIM) to assess disparity map quality;

        \item Empirical evidence demonstrating that GA-optimized parameters significantly
            outperform manually tuned configurations, particularly when dealing
            with varied imaging conditions; and

        \item A practical solution for resource-constrained UAV systems that achieves
            a balance between processing speed and depth estimation accuracy for
            UAV branch pruning applications.
    \end{itemize}

    \section{Related Work}

    \subsection{Stereo Matching}
    Stereo matching \cite{scharstein2002taxonomy} estimates depth by finding corresponding
    points in images from different viewpoints, generating disparity maps
    transformable into depth maps. Disparity estimation precision directly impacts
    depth measurement accuracy, crucial for UAVs balancing accuracy and
    processing speed.

\begin{figure}[t]
        \centering
        \includegraphics[width=1\columnwidth]{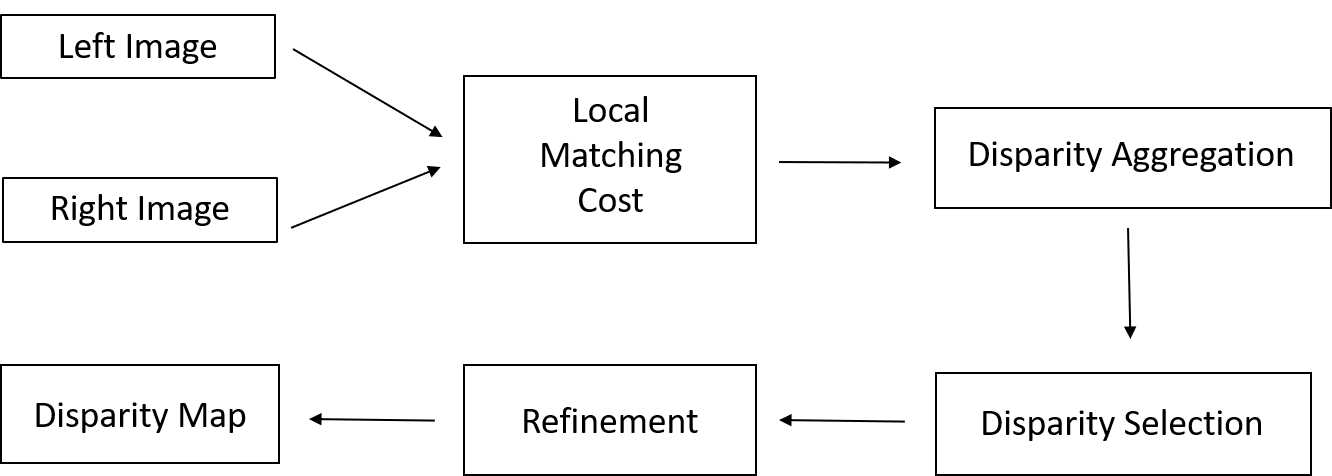}
        \caption{Flowchart of traditional disparity map generation process.}
        \label{traditional_flowchart}
    \end{figure}

    Traditional methods range from simple Block Matching (BM)
    \cite{brown1992survey} using Sum of Absolute Differences (SAD)
    \cite{hirschmuller2007stereo} or Sum of Squared Differences (SSD)
    \cite{scharstein2002taxonomy} to advanced deep learning like PSMNet
    \cite{chang2018pyramid} and RAFT-Stereo \cite{lipson2021raft}. Fig.~\ref{traditional_flowchart}
    shows a typical pipeline. While neural networks offer higher accuracy, their
    processing times (over 1s/frame) are prohibitive for real-time UAVs, unlike optimized
    traditional methods (approx. 0.5s/frame).

    The disparity-depth relationship \cite{hamzah2016literature} is:
    \begin{equation}
        Z = \frac{f \cdot B}{d}\label{eq:depth_disparity}
    \end{equation}
    where $Z$ is depth, $f$ focal length, $B$ baseline, and $d$ disparity. Small
    disparity errors cause large depth errors, especially for distant objects.
    Our approach uses deep learning outputs as benchmarks to optimize SGBM
    parameters, balancing accuracy and efficiency for UAV operations.
    \subsection{Semi-Global Block Matching and Weighted Least Squares for
    Disparity Map Generation}
    SGBM \cite{hirschmuller2007stereo}, an extension of Semi-Global Matching (SGM)
    \cite{hirschmuller2005accurate}, incorporates block-based matching and
    approximates global optimization via multi-directional cost aggregation, balancing
    efficiency and accuracy.

\begin{figure}[t]
        \centering
        \includegraphics[scale=0.35]{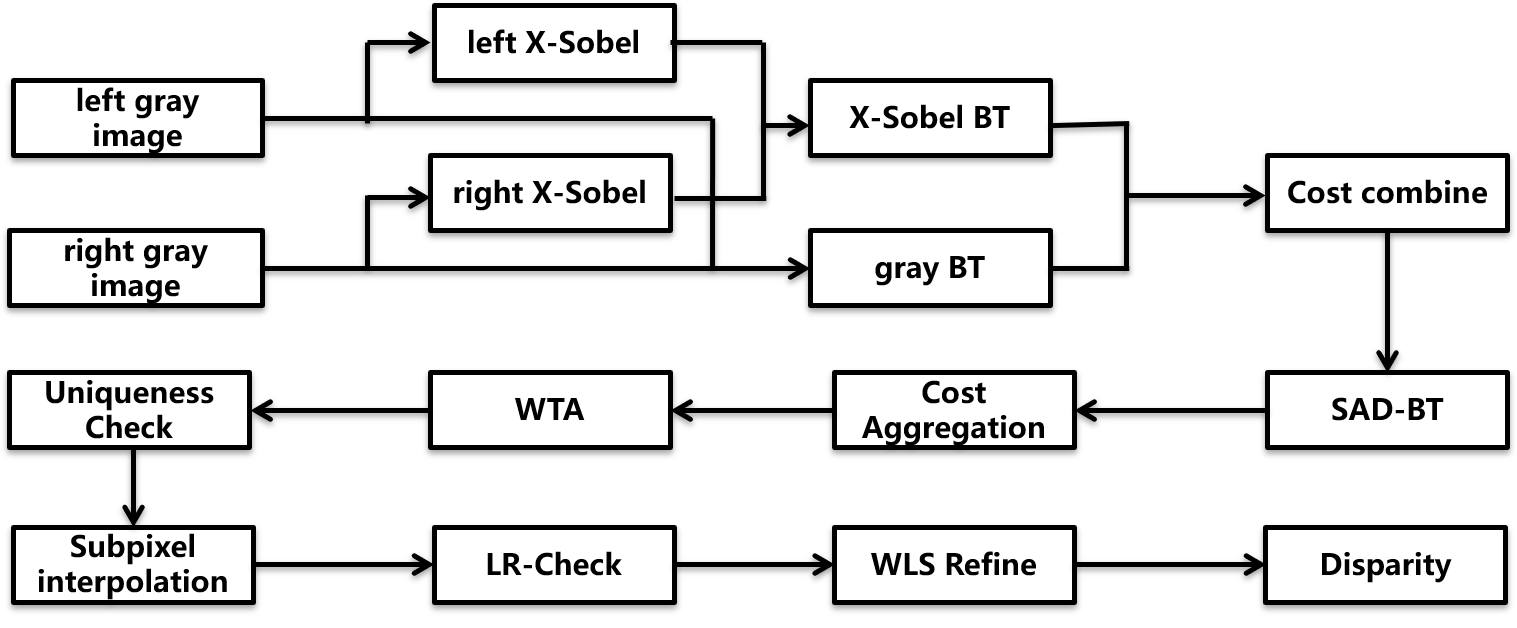}
        \caption{SGBM and WLS processing pipeline for disparity map generation.}
        \label{sgbm_flow_chart}
    \end{figure}

    SGBM steps: (1) Compute initial matching costs using block-based intensity
    differences. (2) Aggregate costs along multiple paths for smoothness. (3)
    Select optimal disparities (minimum aggregated cost). (4) Post-process (sub-pixel
    interpolation, consistency checks, outlier removal). We further apply WLS
    filtering to preserve edges and smooth homogeneous regions. Fig.~\ref{sgbm_flow_chart}
    illustrates this pipeline.

    The SGBM energy function is:
    \begin{equation}
        E(D) = \sum_{p}C(p, d_{p}) + \sum_{p, q \in N(p)}P(d_{p}, d_{q}) \label{eq:sgbm_energy}
    \end{equation}
    where $C(p, d_{p})$ is the matching cost, usually measured by intensity or
    feature differences, and $P(d_{p}, d_{q})$ penalizes disparity differences.

    Cost aggregation along path $r$:
    \begin{equation}
        \begin{aligned}
            L_{r}(p,d) = C(p,d) & + \min\begin{cases}L_{r}(p-r,d) \\ L_{r}(p-r,d\pm1) + \alpha \\ \min\limits_{k}L_{r}(p-r,k) + \beta\end{cases} \\
                                & - \min\limits_{k}L_{r}(p-r,k)
        \end{aligned}
        \label{eq:aggregation}
    \end{equation}

    WLS refinement:
    \begin{equation}
        E_{\text{WLS}}(D)=\sum_{p}(D(p)-\tilde{D}(p))^{2}+\lambda\sum_{(p,q)\in N}
        w_{pq}(D(p)-D(q))^{2}\label{eq:wls_energy}
    \end{equation}
    where $\tilde{D}$ is the initial disparity map, $\lambda$ regularization strength,
    and $w_{pq}$ edge-aware weights.

    The pipeline starts with preprocessing (grayscale conversion, Sobel filter).
    Birchfield-Tomasi matching cost for pixel $p=(x,y)$ and disparity $d$ (with $q
    =(x-d,y)$):
    \begin{align}
        I_{\max}(p,d)        & = \max\{I_{r}(q-1), I_{r}(q), I_{r}(q+1)\}                               \\
        I_{\min}(p,d)        & = \min\{I_{r}(q-1), I_{r}(q), I_{r}(q+1)\}                               \\
        C_{\text{gray}}(p,d) & = \max\left\{0,\;I_{l}(p)-I_{\max}(p,d),\;I_{\min}(p,d)-I_{l}(p)\right\}
    \end{align}
    Gradient costs ($G_{l}, G_{r}$ are gradient magnitudes):
    \begin{align}
        G_{\max}(p,d)        & = \max\bigl\{G_{r}(q-1),\; G_{r}(q),\; G_{r}(q+1)\bigr\}                                                   \\
        G_{\min}(p,d)        & = \min\bigl\{G_{r}(q-1),\; G_{r}(q),\; G_{r}(q+1)\bigr\}                                                   \\
        C_{\text{grad}}(p,d) & = \max\Bigl\{0,\; \begin{aligned}[t]&G_{l}(p)-G_{\max}(p,d), \\&G_{\min}(p,d)-G_{l}(p)\Bigr\}\end{aligned}
    \end{align}
    Combined cost:
    \begin{equation}
        C(p,d) = \frac{(100-\eta)C_{\text{gray}}(p,d) + \eta C_{\text{grad}}(p,d)}{100}
        , \quad 1 \leq \eta \leq 100
    \end{equation}
    SGM \cite{hirschmuller2005accurate} aggregates costs along multiple paths $r$
    (Eq. \ref{eq:aggregation}). Aggregated cost: $S(p,d) = \sum_{r}L_{r}(p,d)$. Initial
    disparity: $d^{*}(p) = \arg\min_{d}S(p,d)$. Sub-pixel refinement:
    \begin{equation}
        d_{\text{sub}}(p) = d^{*}(p) + \frac{S(d^{*}-1) - S(d^{*}+1)}{2[S(d^{*}-1)
        + S(d^{*}+1) - 2S(d^{*})]}
    \end{equation}
    Post-processing: Uniqueness check ($S_{2}- S_{1}\geq \frac{\gamma}{100}S_{1}$),
    left-right consistency ($|D_{L}(x,y)-D_{R}(x-D_{L}(x,y),y)| \leq \Delta$), speckle
    filtering ($|R| < W \;\land\; \max_{p,q\in R}|D_{L}(p)-D_{L}(q)| \leq \delta$).
    WLS refinement (Eq. \ref{eq:wls_energy}) uses edge-aware weights $w_{pq}=\exp
    \left(-\frac{\|I(p)-I(q)\|^{2}}{2\sigma^{2}}\right)$. GA-optimized SGBM and
    WLS provide accurate, efficient disparity estimation for UAVs.

    \subsection{Genetic Algorithms for Parameter Optimization}
    GAs \cite{ming2024constrained}, inspired by natural selection, are powerful optimizers,
    widely used in scheduling, machine learning, and multi-objective
    optimization. They are effective for optimizing stereo vision algorithm parameters.

    GA's main advantage is effectively exploring complex, non-linear parameter
    spaces. Traditional SGBM/WLS parameter tuning is manual, time-consuming, and
    often suboptimal. GAs systematically evaluate parameter combinations to
    maximize disparity map quality.

    The GA process have several steps: initialization (candidate parameter sets),
    fitness evaluation (MSE \cite{sara2019image}, PSNR \cite{sara2019image}, SSIM
    \cite{wang2004image}), selection, crossover, and mutation. This automates
    SGBM/WLS parameter optimization, finding configurations outperforming manual
    tuning, especially against high-quality gold standard from RAFT-Stereo with
    NeRF-Supervision.

    \section{The Proposed Method}
    Despite its efficiency, traditional SGBM implementations face significant challenges
    in parameter selection, which directly impacts disparity map quality. Recognizing
    these limitations, we leverage advanced optimization techniques to overcome
    these challenges. In this study, we employ GA \cite{ming2024constrained} to
    automatically tune the SGBM parameters, thereby enhancing the accuracy of disparity
    map estimation. Furthermore, we apply WLS filtering \cite{romano2017resurrecting}
    to improve the smoothness and reliability of depth estimates, as parameter
    choices significantly impact the resulting disparity maps.

    In this section, we detail the key parameters of SGBM and WLS, and describe
    the encoding process used for their optimization. We also outline the methodology
    for employing a GA to search for the optimal parameter configuration. The
    evaluation of these optimized parameters is presented in Section IV.

    \subsection{Parameters of SGBM and WLS}
    Our stereo disparity estimation method combines the SGBM
    \cite{hirschmuller2007stereo} approach with a WLS filter
    \cite{farbman2008edge} to produce a refined disparity map. As demonstrated in
    the preceding mathematical formulations, the quality of the resulting
    disparity map is highly sensitive to multiple parameter settings that
    control different aspects of the stereo matching pipeline. These parameters directly
    influence the energy terms in Equations (\ref{eq:sgbm_energy}), (\ref{eq:aggregation}),
    and (\ref{eq:wls_energy}), creating a complex parameter space with
    interdependent effects on disparity estimation accuracy.

    The smoothness penalties $\alpha$ and $\beta$ in Equation (\ref{eq:aggregation})
    critically determine how disparity discontinuities are handled during cost
    aggregation. A small penalty $\alpha$ leads to minor disparity changes (typically
    at slanted surfaces), while a larger penalty $\beta$ affects significant
    disparity jumps (usually at object boundaries). Optimizing the relationship between
    these penalties directly impacts the preservation of depth discontinuities and
    smoothness of homogeneous regions. Similarly, the gradient-intensity weight $\eta$
    balances the contribution of intensity-based and gradient-based matching costs,
    affecting the algorithm's robustness to illumination variations while
    preserving structural details.

    Post-processing parameters including the left-right consistency threshold
    $\Delta$, uniqueness ratio $\gamma$, and speckle filtering parameters ($W$ and
    $\delta$) govern the detection and handling of occlusions, ambiguous matches,
    and noise. Finally, the WLS parameters $\lambda$ and $\sigma$ control the
    edge-preserving smoothing operation described in Equation (\ref{eq:wls_energy}),
    with $\lambda$ determining regularization strength and $\sigma$ adjusting edge
    sensitivity.

    Table \ref{table_param_summary} summarizes these nine critical parameters along
    with their valid ranges. Each parameter changes within specific constraints
    and interacts with other parameters in complex, nonlinear ways that are difficult
    to model analytically. This complexity makes manual parameter tuning
    exceptionally challenging and motivates our GA-based optimization approach
    as described in the following section.

    \begin{table}[t]
        \centering
        \caption{Summary of Parameters for SGBM and WLS Optimization}
        \label{table_param_summary}
        \begin{tabular}{cll}
            \hline
            Parameter & Description               & Valid Range                           \\
            \hline
            $\alpha$  & Small smoothness penalty  & $1 \leq \alpha \leq 100000$           \\
            $\beta$   & Large smoothness penalty  & $\alpha < \beta \leq 200000$          \\
            $\Delta$  & LR consistency threshold  & $1 \leq \Delta \leq 100$              \\
            $\eta$    & Gradient-intensity weight & $1 \leq \eta \leq 100$                \\
            $\gamma$  & Uniqueness ratio          & $1 \leq \gamma \leq 100$              \\
            $W$       & Speckle window size       & $1 \leq W \leq 1000$                  \\
            $\delta$  & Speckle range             & $1 \leq \delta \leq 1000$             \\
            $\lambda$ & WLS regularization        & $1 \leq \lambda \leq 100000$          \\
            $\sigma$  & WLS edge sensitivity      & $0 \leq \sigma \leq 0.99$ (step 0.01) \\
            \hline
        \end{tabular}
    \end{table}

    \subsection{Parameter Encoding for GA Optimization}
    In the previous subsection, we introduced the key parameters governing the
    SGBM and WLS algorithms, outlining their definitions and valid value ranges.
    In this section, we describe how a real code GA is utilized to optimize these
    parameters. It is important to note that even minor adjustments to these
    parameters within the SGBM and WLS frameworks can substantially impact the
    quality of the resulting disparity map.

    To optimize the nine parameters using a GA, each is encoded as an integer variable
    with a range from 1 to 10 in Table~\ref{table_param_summary}. 29 integer
    variables, $x_{1}$, $x_{2}$, ..., $x_{9}$ are used to reprensent the nine
    parameters. Each parameter is represented by one or more of these integer variables.
    For example, since the range of $\alpha$ is $1 \leq \alpha \leq 100\,000$,
    it is computed using five integer genes, $x_{1}$, $x_{2}$, $x_{3}$, $x_{4}$,
    and $x_{5}$, as follows:
    \begin{equation}
        \begin{split}
            \alpha = \text{int}\Bigl(&(x_{1}- 1) \times 10000 + (x_{2}- 1) \times
            1000 \\
            &+ (x_{3}- 1) \times 100 + (x_{4}- 1) \times 10 + x_{5}\Bigr ) ,
        \end{split}
    \end{equation}
    where $x_{1}, x_{2}, x_{3}, x_{4}, x_{5}\in [1,10]$. Similarly, other parameters
    are encoded using their respective variables: $\beta$ uses $x_{6}$ through
    $x_{10}$, $\Delta$ uses $x_{11}$ and $x_{12}$, $\eta$ uses $x_{13}$ and
    $x_{14}$, $\gamma$ uses $x_{15}$ and $x_{16}$, $W$ uses $x_{17}$ through
    $x_{19}$, $\delta$ uses $x_{20}$ through $x_{22}$, $\lambda$ uses $x_{23}$
    through $x_{27}$, and $\sigma$ uses $x_{29}$.

\begin{figure*}[t] 
        \centering
        \includegraphics[width=1\textwidth]{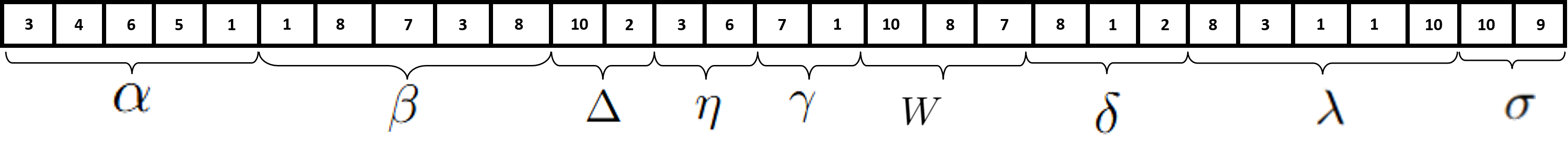} 
        \caption{Parameter encoding example for GA optimization of SGBM and WLS
        parameters, which is a 29-bit integer vector.}
        \label{encoding}
    \end{figure*}

    In the GA method, a 29-bit vector is used to represent the nine parameters
    in the SGBM and WLS algorithms as chromosome. Each bit ranges from 1 to 10. This
    encoding ensures a uniform representation of the entire search space, as
    illustrated in Fig.~\ref{encoding}.

    Based on the above encoding/representation methodm, GA utilizes a custom two-point
    crossover operator that randomly selects two crossover points along the
    chromosome and exchanges the segment between them from two parent
    individuals. This operation effectively combines advantages from both parents
    while ensuring that all gene values remain within the prescribed integer bounds—each
    gene is rounded and constrained within the range.

    To further enhance exploration of the parameter space and prevent premature convergence,
    a uniform integer mutation operator is applied. With a mutation probability of
    0.3 for each gene, any mutated gene is replaced by a new integer randomly
    chosen from the range. This mechanism introduces the necessary randomness to
    maintain genetic diversity across the population.

    Following the generation of offspring through crossover and mutation, an
    elitist selection strategy is employed. Specifically, the top five
    individuals, based on their fitness scores, are retained for the next
    generation.

    \begin{figure}[t]
        \centering
        \includegraphics[width=0.5\textwidth]{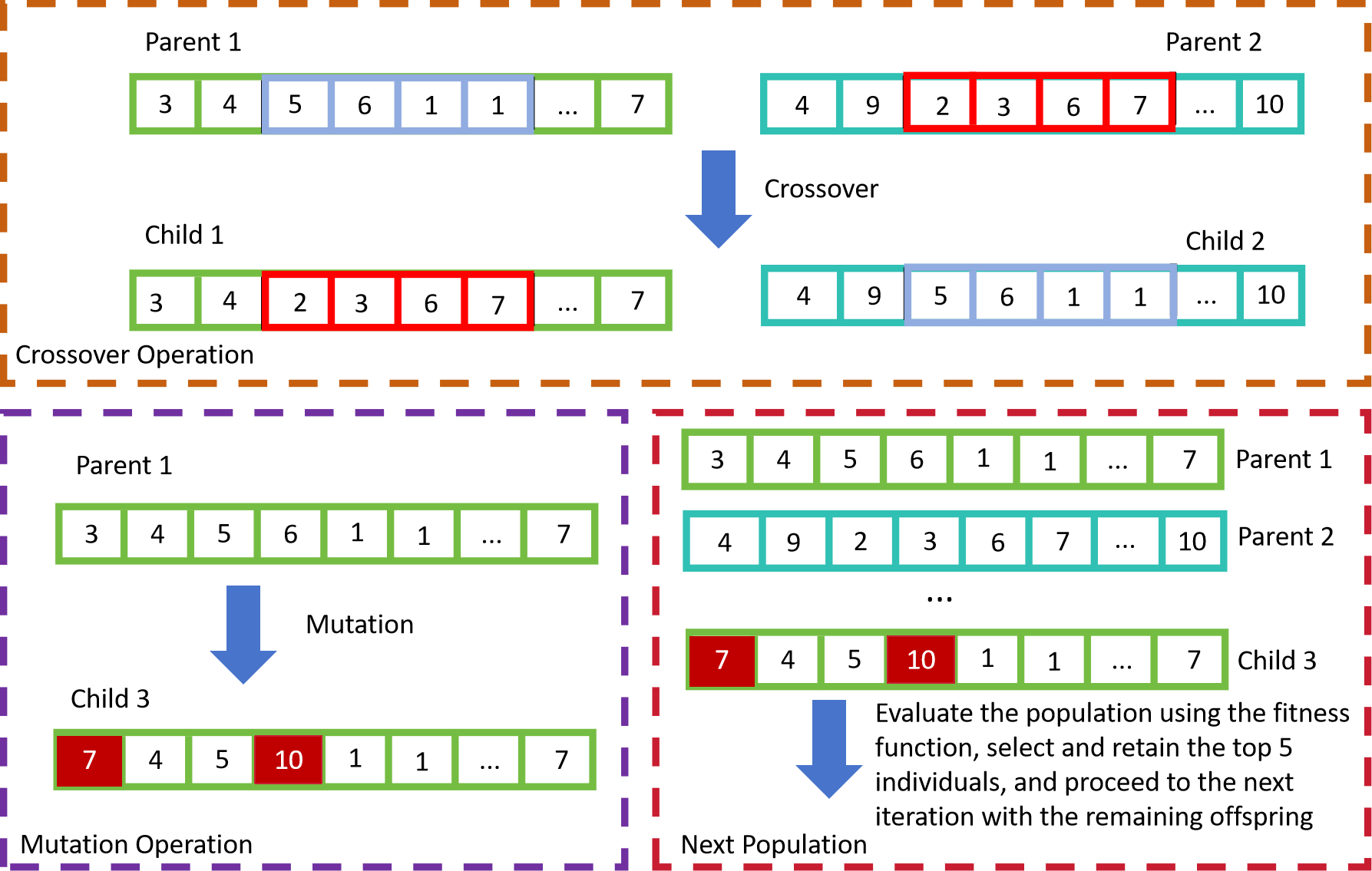}
        \caption{Crossover, mutation, and environment selection in GA.}
        \label{GA_process}
    \end{figure}
    \subsection{Fitness Evaluation}
    For fitness evaluation within our GA framework, we employe three different image
    quality metrics: MSE \cite{sara2019image}, PSNR \cite{sara2019image}, and
    SSIM \cite{wang2004image}. Each metric served as an independent fitness functions,which
    forms three variants of the proposed GA methods. This allowing us to analyze
    how different evaluation criteria influence parameter selection and
    resulting disparity map characteristics.

    \begin{equation}
        MSE = \frac{1}{N}\sum_{i=1}^{N}\bigl(D_{GT}(i) - D_{SGBM+WLS}(i)\bigr)^{2}
    \end{equation}

    MSE quantifies pixel-level accuracy by measuring the squared difference between
    corresponding pixels in the ground truth disparity map ($D_{GT}$) and our SGBM+WLS
    generated disparity map ($D_{SGBM+WLS}$). As a fitness function, it drives the
    GA search toward parameter sets that minimize overall numerical error. The
    squaring operation makes MSE particularly sensitive to large disparities, encouraging
    the identification of parameters that avoid significant depth estimation errors
    critical for collision avoidance in UAV navigation. However, MSE treats all pixels
    with equal importance regardless of their visual significance or structural context.

    \begin{equation}
        PSNR = 10 \log_{10}\left(\frac{d_{\max}^{2}}{MSE}\right)
    \end{equation}

    PSNR expresses quality on a logarithmic scale relative to the maximum
    possible disparity value ($d_{\max}$). When used as a fitness function, it
    tends to produce parameter configurations that achieve more balanced error
    distributions across the disparity map. The decibel (dB) scale provides intuitive
    interpretation, with higher values indicating better quality and each 6dB
    increase approximately corresponding to a doubling of signal quality. PSNR-guided
    optimization typically results in parameters that produce more consistent depth
    representation across the entire image, with particular improvements in
    homogeneous regions.

    \begin{equation}
        \begin{split}
            &SSIM(D_{GT}, D_{SGBM+WLS}) = \\
            &\frac{(2\mu_{GT}\mu_{SGBM+WLS}+C_{1})(2\sigma_{GT,SGBM+WLS}+C_{2})}{(\mu_{GT}^{2}+\mu_{SGBM+WLS}^{2}+C_{1})(\sigma_{GT}^{2}+\sigma_{SGBM+WLS}^{2}+C_{2})}
        \end{split}
    \end{equation}

    SSIM evaluates structural similarity rather than pixel-wise differences,
    incorporating luminance ($\mu$), contrast ($\sigma$), and structure ($\sigma_{GT,SGBM+WLS}$)
    components. This perceptually aligned metric ranges from -1 to 1, with 1 indicating
    perfect structural similarity. When employed as a fitness function, SSIM guides
    the GA toward parameter configurations that preserve edge structures and textural
    patterns, particularly important for maintaining the integrity of tree
    branch boundaries and depth discontinuities. Our experimental analysis showed
    that while MSE optimization minimized overall error, PSNR balanced signal
    quality with region consistency, and SSIM best preserved structural fidelity—making
    SSIM-optimized parameters most suitable for UAV pruning, as detailed in
    Section V.

    \section{Results and Analysis}
    This section systematically evaluates our GA-based parameter optimization
    for SGBM and WLS, detailing gold standard disparity acquisition, metric-based
    analysis, and comparisons with manual tuning to demonstrate robustness and generalizability.

    \subsection{Gold standard Acquisition using NeRF-Supervised RAFT-Stereo}
\begin{figure}[t]
        \centering
        \resizebox{0.5\textwidth}{!}{ 
        \begin{tabular}{cc}
            \begin{subfigure}{0.45\textwidth} \centering \includegraphics[width=\linewidth]{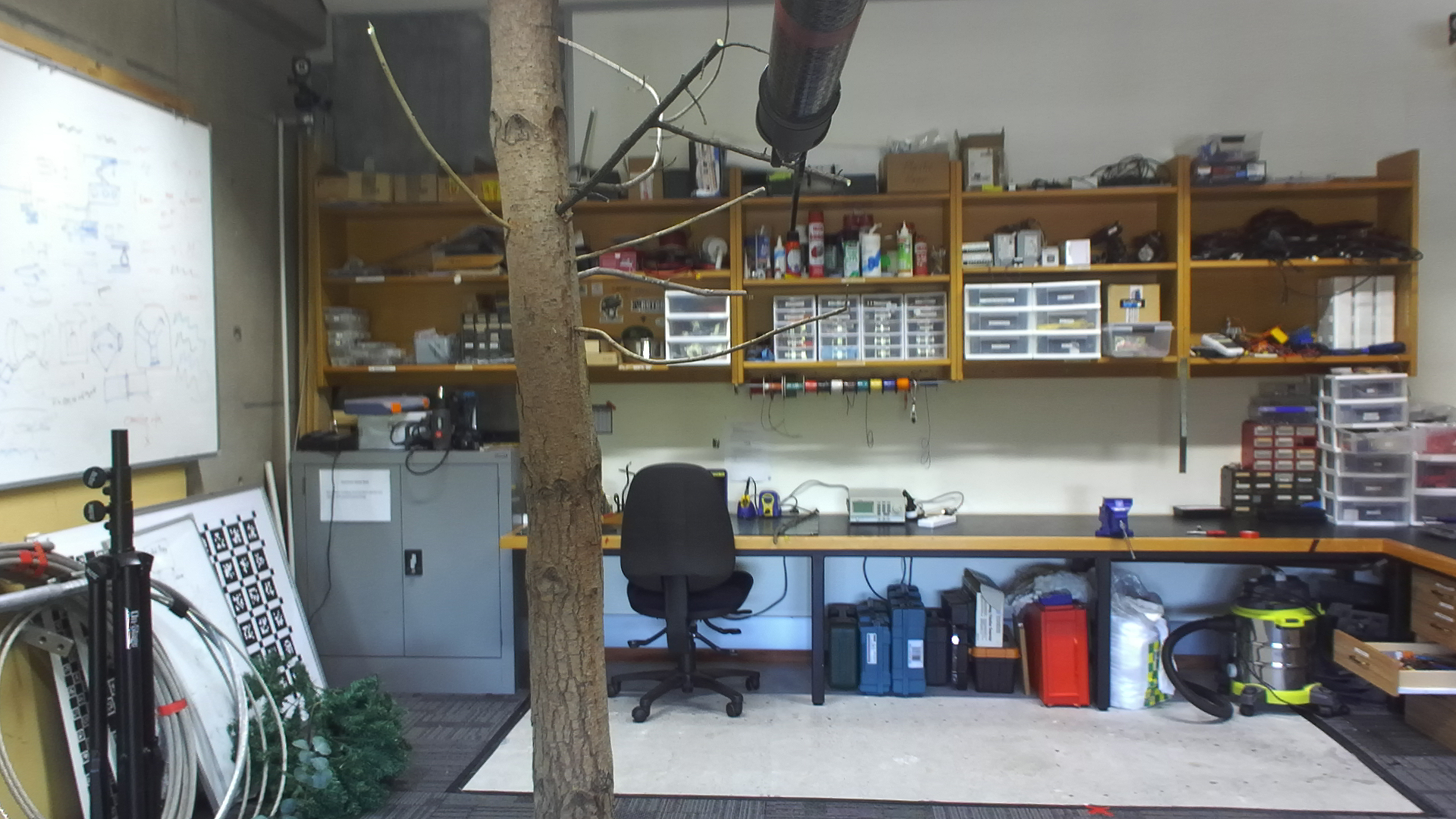} \caption{Left image of Example 33}\end{subfigure}   & \begin{subfigure}{0.45\textwidth} \centering \includegraphics[width=\linewidth]{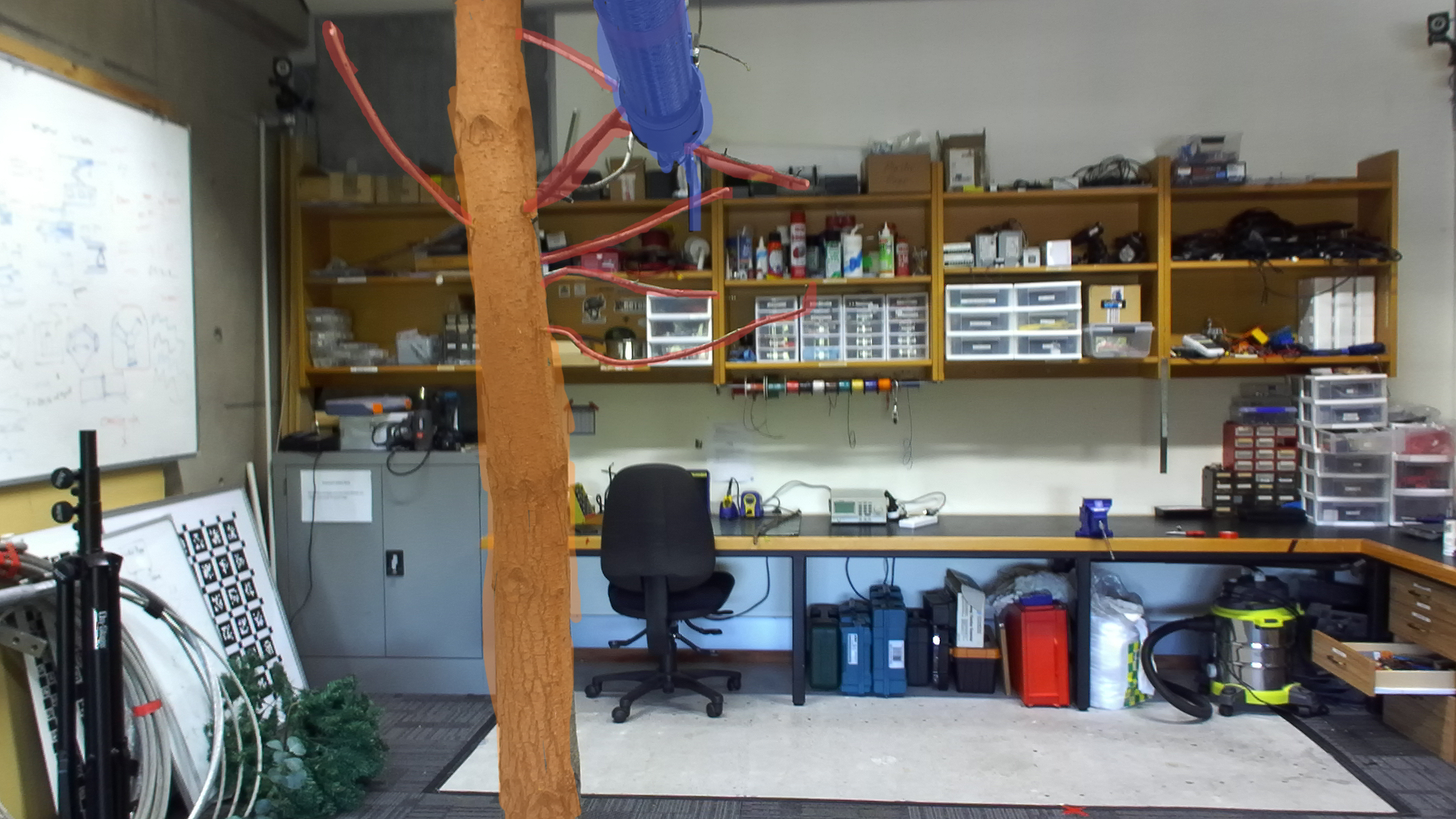} \caption{Right image of Example 33}\end{subfigure}   \\
            \begin{subfigure}{0.45\textwidth} \centering \includegraphics[width=\linewidth]{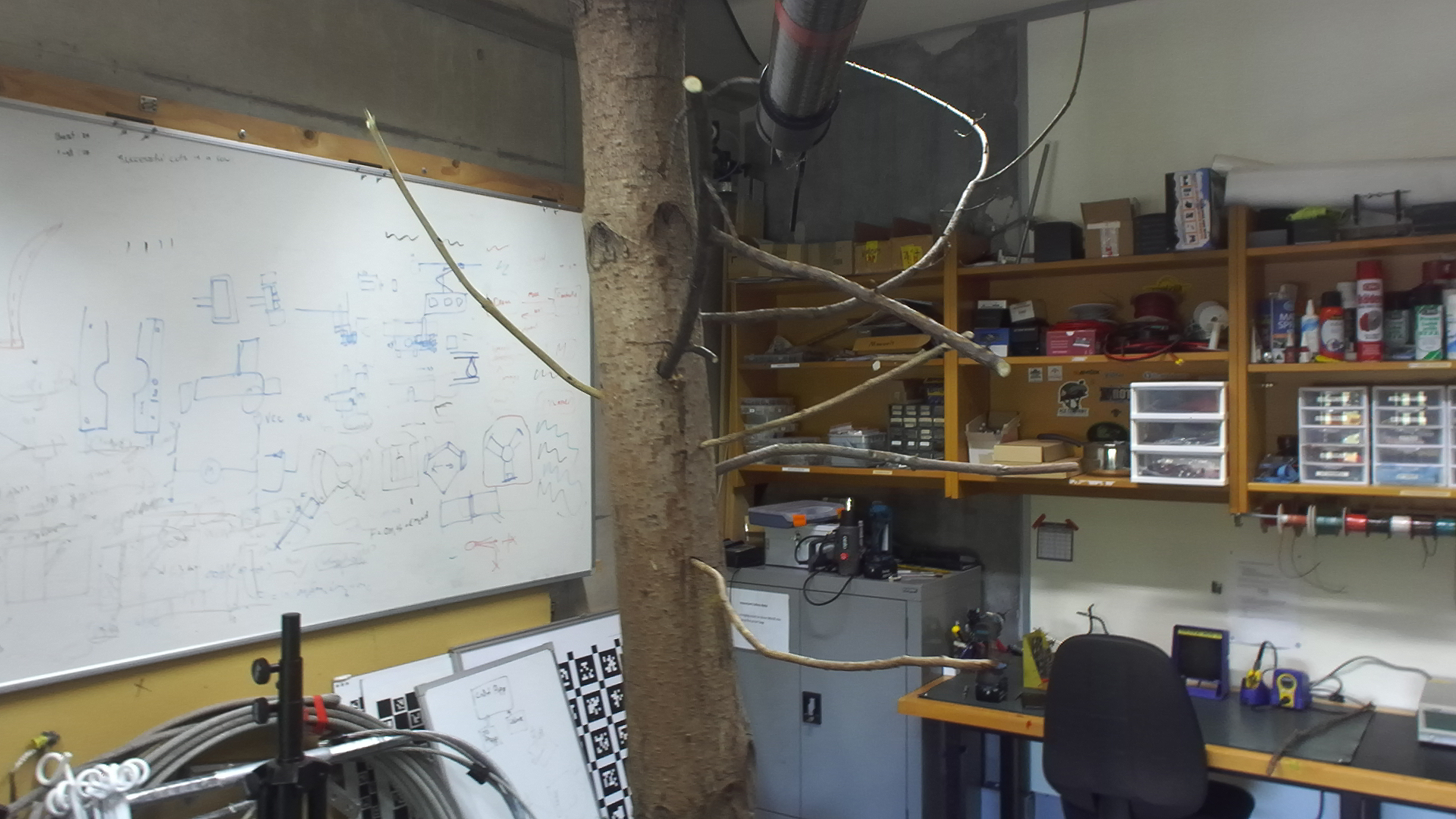} \caption{Left image of Example 474}\end{subfigure} & \begin{subfigure}{0.45\textwidth} \centering \includegraphics[width=\linewidth]{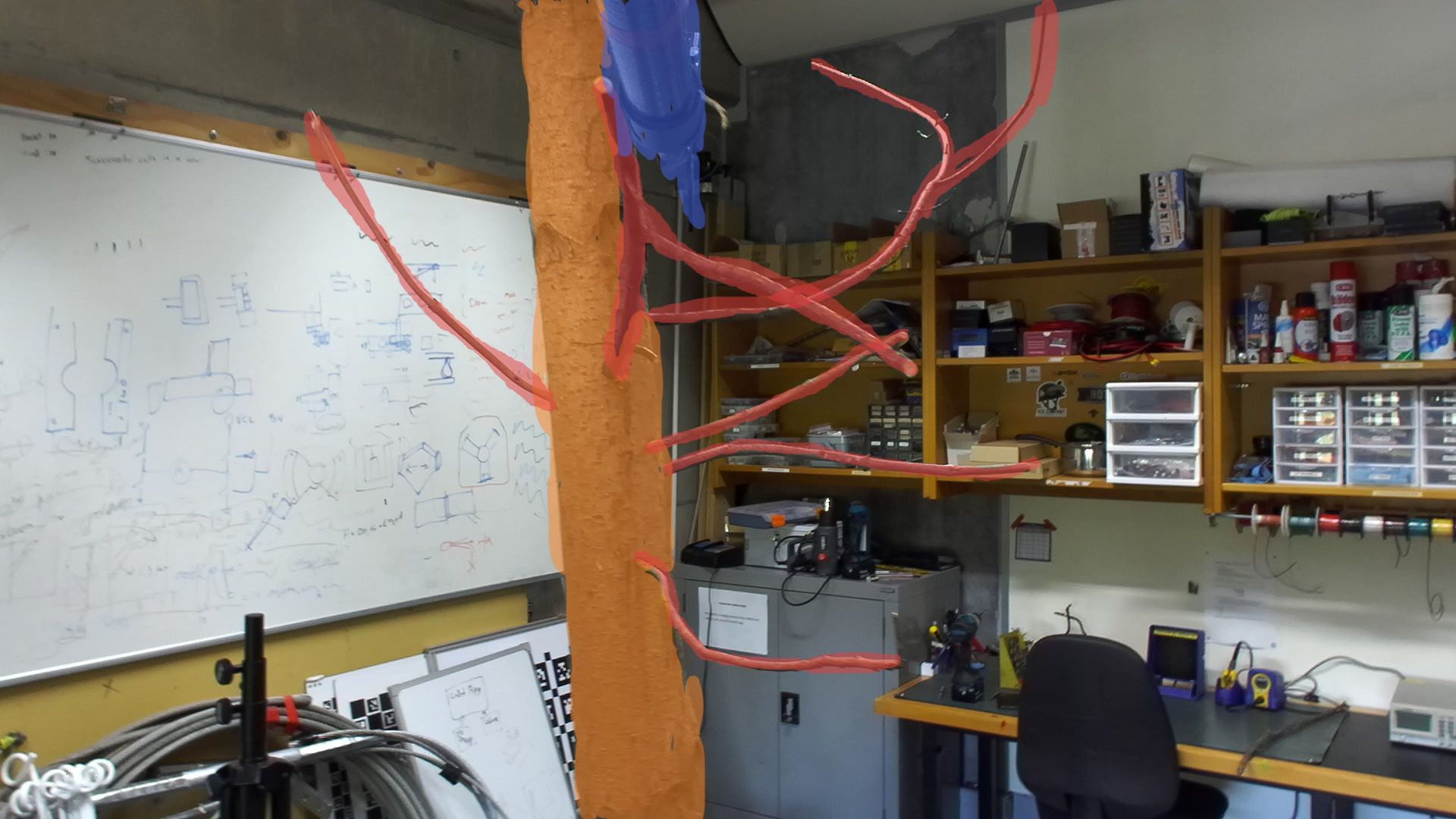} \caption{Right image of Example 474}\end{subfigure}
        \end{tabular}
        }
        \caption{Left and right stereo image pairs captured using a ZED Mini
        camera mounted on a UAV in a Lab environment. Blue markings indicate UAV
        pruning tools, red markings highlight tree branches, and orange markings
        identify tree trunks.}
        \label{nerf_ground_truth}
    \end{figure}

\begin{figure}[t]
        \centering
        \resizebox{0.5\textwidth}{!}{ 
        \begin{minipage}{\textwidth}
            \centering
            \begin{subfigure}
                {0.40\textwidth}
                \includegraphics[width=\linewidth]{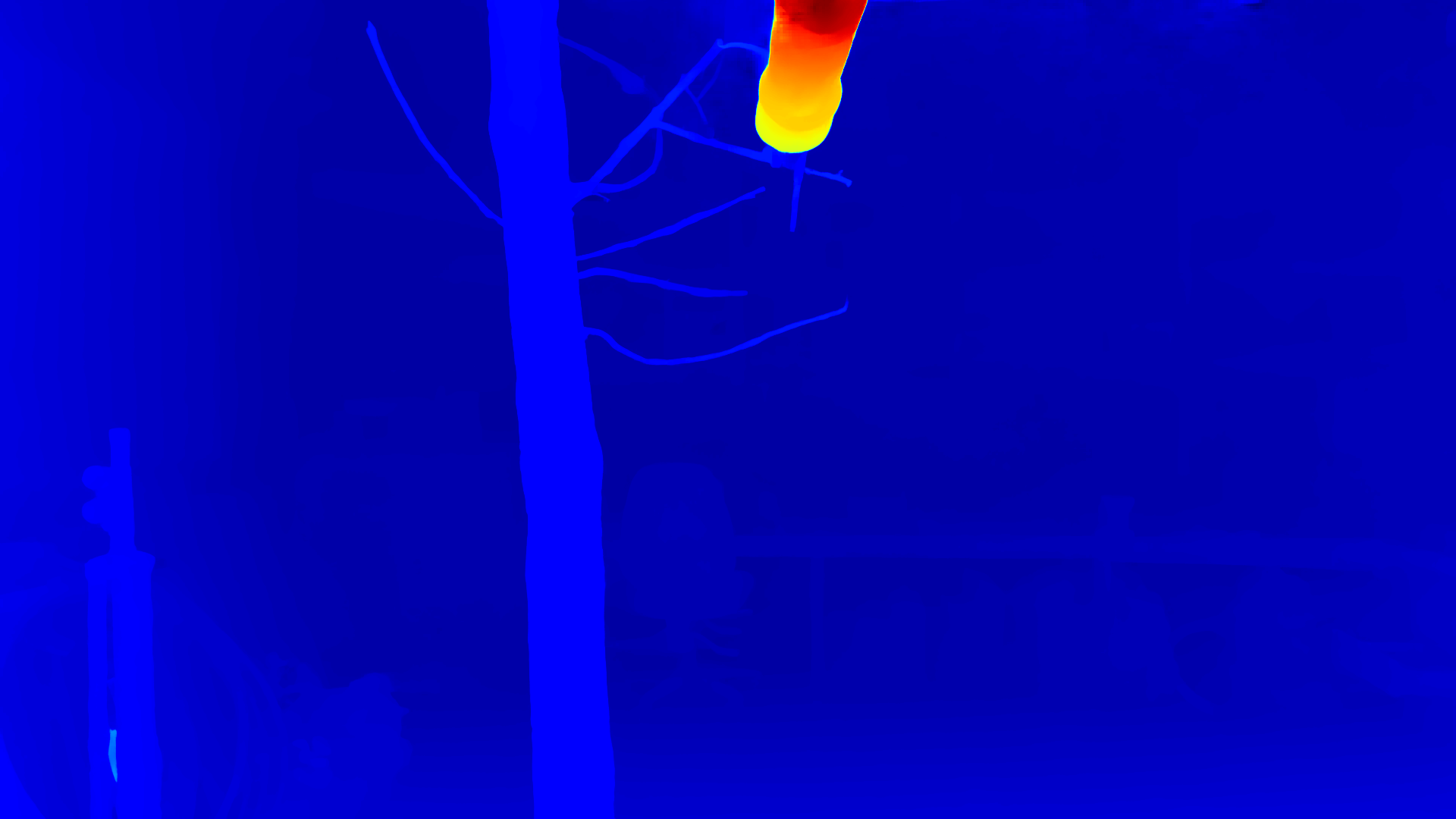}
                \caption{Gold standard disparity map of Example 33}
            \end{subfigure}\hspace{0.1\textwidth}
            \begin{subfigure}
                {0.40\textwidth}
                \includegraphics[width=\linewidth]{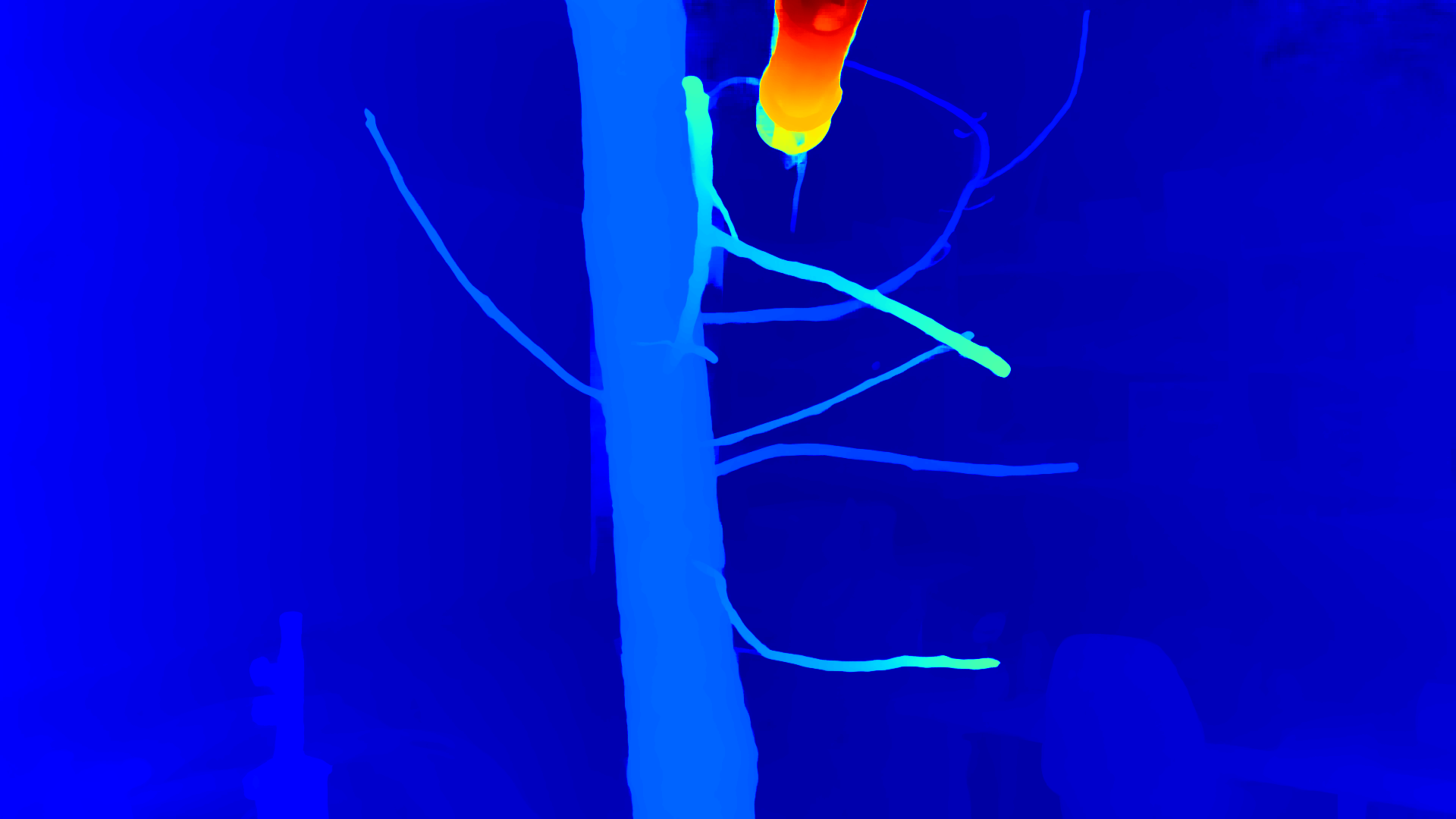}
                \caption{Gold standard disparity map of Example 474}
            \end{subfigure}
        \end{minipage}
        }
        \caption{Reference disparity maps generated using RAFT-Stereo with NeRF-Supervised
        training for quantitative evaluation.}
        \label{nerf_disparity_map}
    \end{figure}

    We created high-quality disparity maps using NeRF-Supervised \cite{tosi2023nerf}
    RAFT-Stereo \cite{lipson2021raft} as gold standard for evaluation. A ZED
    Mini stereo camera mounted on a UAV captured 654 high-definition (1920×1080)
    RGB stereo image pairs of Radiata Pine branches under consistent lighting
    conditions. Gold standard disparity maps were generated by training RAFT-Stereo
    on KITTI2012 \cite{geiger2012we} and KITTI2015 \cite{Menze2015CVPR} datasets,
    fine-tuning on our forestry dataset, and refining with NeRF-Supervised loss.
    This yielded detailed disparity maps accurately delineating tree branch
    structures. For systematic evaluation, we selected two representative test
    cases—Examples 33 and 474—containing varying branch structures and imaging
    conditions. Fig.~\ref{nerf_ground_truth} presents the original left and
    right RGB images of these examples, while Fig.~\ref{nerf_disparity_map}
    displays their corresponding gold standard disparity maps.

    \subsection{GA-Based Parameter Optimization}
    We applied our GA framework to Example 33 with parameters shown in Table~\ref{GA_parameters}.
    We used 30 population size and 100 generations to balance exploration and efficiency,
    with crossover and mutation probabilities of 0.6 and 0.3 respectively.
    \begin{table}[t]
        \centering
        \caption{Parameters of the GA Optimizer}
        \label{GA_parameters}
        \begin{tabular}{@{}ll@{\hspace{1.5cm}}ll@{}}
            \hline
            \textbf{Parameter Name} & \textbf{Value} & \textbf{Parameter Name} & \textbf{Value} \\
            \hline
            Population size         & 30             & Crossover probability   & 0.6            \\
            Chromosome length       & 29             & Mutation probability    & 0.3            \\
            Number of generations   & 100            & Retain top chromosomes  & 5              \\
            \hline
        \end{tabular}
    \end{table}
    To ensure statistical validity, we conducted 30 independent runs for each
    fitness function (MSE, PSNR, SSIM), mitigating GA's stochastic nature and providing
    robust performance assessment.
    \subsection{Quantitative Performance Analysis}
\begin{figure}[htbp]
        \centering
        \includegraphics[width=1\linewidth]{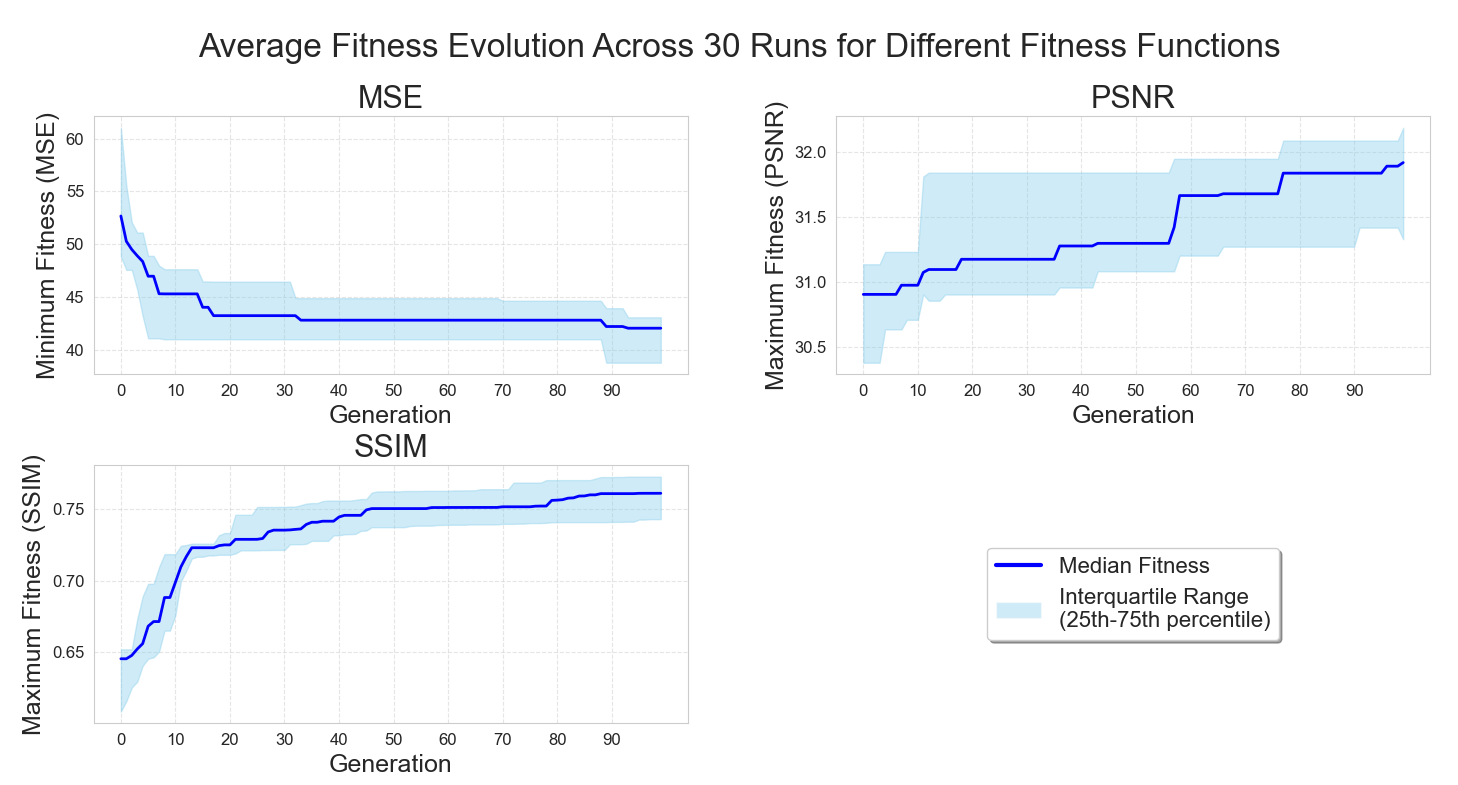}
        \caption{Evolution of evaluation metrics during GA optimization for
        Example 33. Each plot shows average performance over the 30 independent
        runs, with shaded regions indicating standard deviation.}
        \label{example_33_GA}
    \end{figure}

    Fig.~\ref{example_33_GA} illustrates the progression of our three evaluation
    metrics across 100 generations of GA optimization for Example 33. The plots demonstrate
    clear convergence patterns for all metrics, with substantial improvements
    during the first 40 generations followed by more gradual refinements.
    Analyzing convergence trends in Fig.~\ref{example_33_GA}, MSE exhibits rapid
    initial improvement (75\% reduction in first 30 generations), PSNR shows steep
    initial improvement then incremental gains, while SSIM demonstrates
    consistent improvement throughout. Narrowing standard deviation bands confirm
    GA robustness across 30 independent runs. Quantitatively, GA-optimized parameters
    achieved 42.86\% MSE reduction, 8.47\% PSNR increase, and 28.52\% SSIM
    improvement compared to baseline. The substantial SSIM improvement is particularly
    significant as it better captures structural integrity of branch features critical
    for UAV pruning. Parameter configurations optimized for one metric generally
    improve others, indicating alignment between quality aspects while confirming
    each metric emphasizes different disparity map characteristics.

    \subsection{Comparative Analysis on Disparity Map Quality}
\begin{figure}[t]
        \centering
        \resizebox{0.5\textwidth}{!}{ 
        \begin{minipage}{\textwidth}
            \centering
            \begin{subfigure}
                {0.45\textwidth}
                \includegraphics[width=\linewidth]{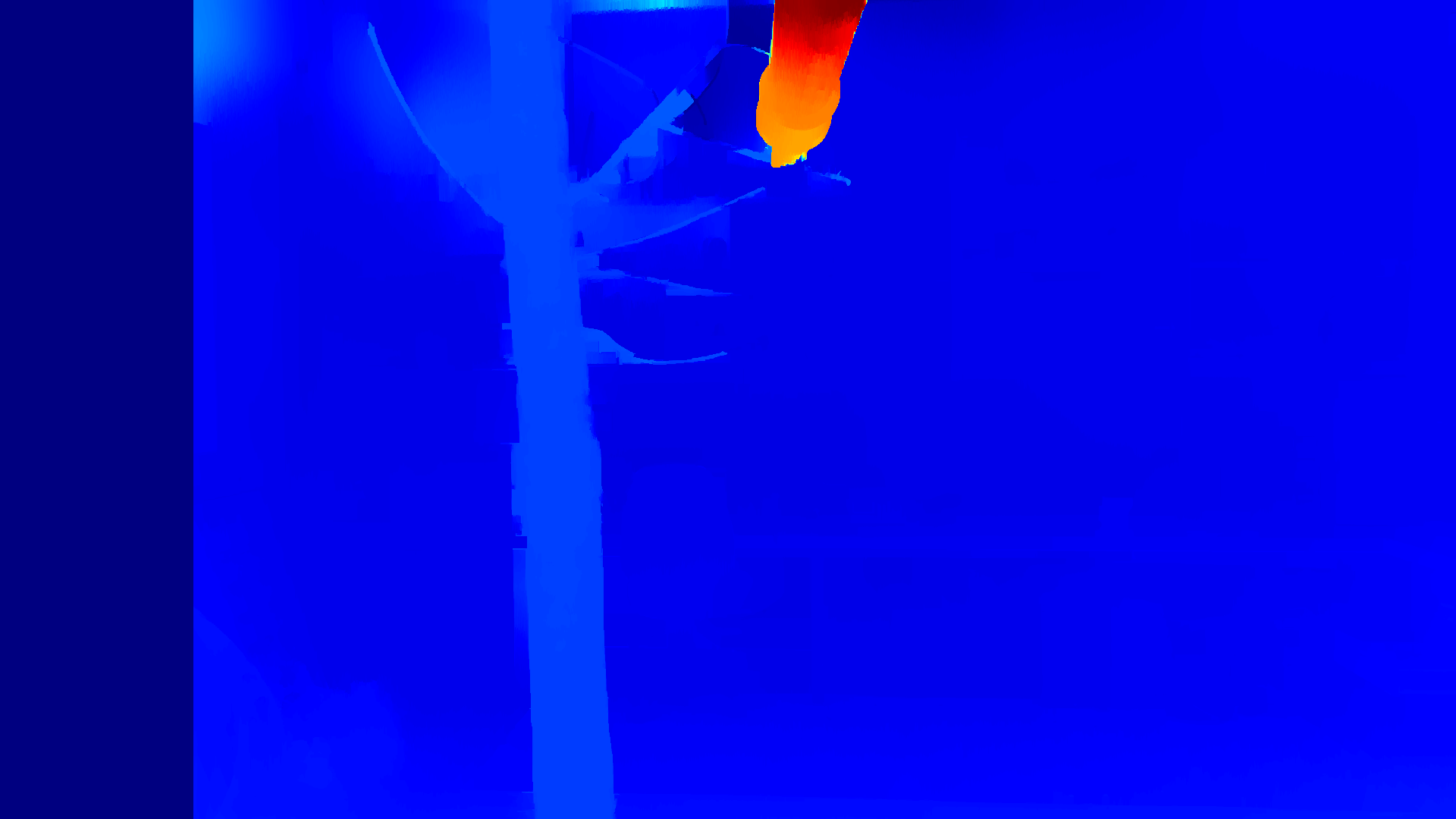}
                \caption{Manual configuration}
            \end{subfigure}\hfill
            \begin{subfigure}
                {0.45\textwidth}
                \includegraphics[width=\linewidth]{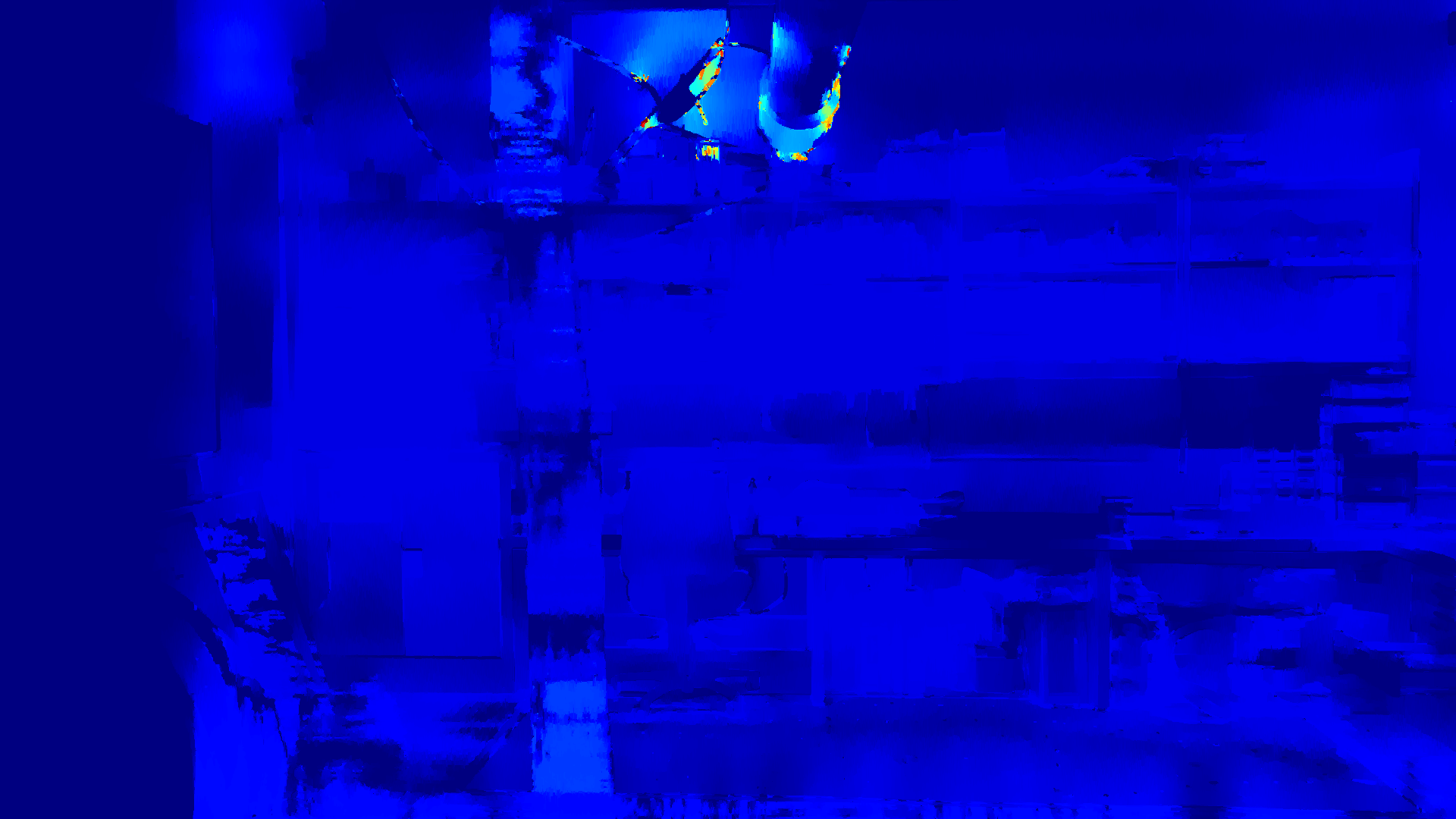}
                \caption{GA optimization with SSIM}
            \end{subfigure}
        \end{minipage}
        }
        \caption{Comparison of disparity maps for Example 33 generated by SGBM+WLS
        using manually configured and GA-optimized parameters.}
        \label{33_example}
    \end{figure}

\begin{figure}[t]
        \centering
        \resizebox{0.5\textwidth}{!}{ 
        \begin{minipage}{\textwidth}
            \centering
            \begin{subfigure}
                {0.45\textwidth}
                \includegraphics[width=\linewidth]{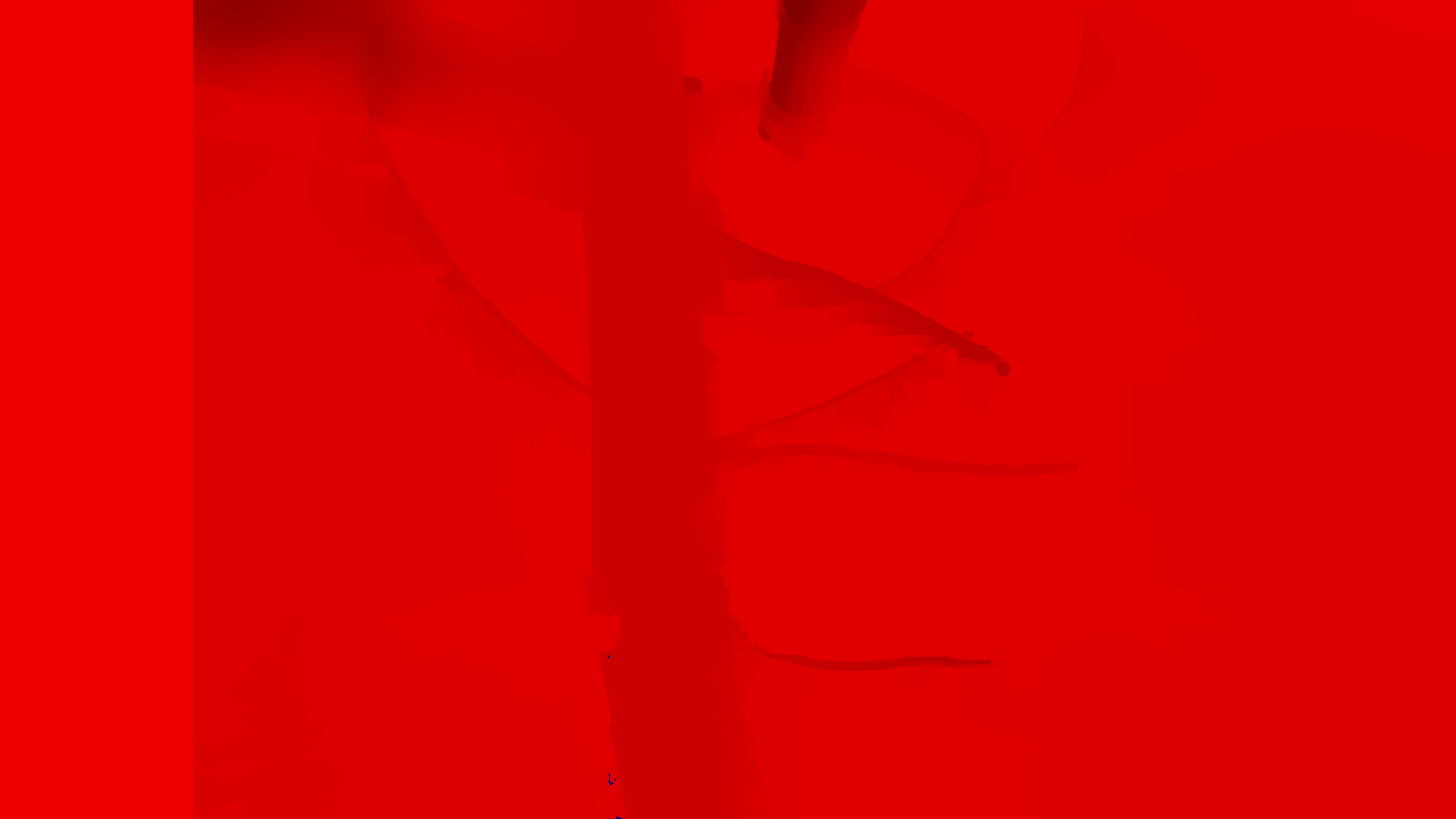}
                \caption{Manual configuration}
            \end{subfigure}\hfill
            \begin{subfigure}
                {0.45\textwidth}
                \includegraphics[width=\linewidth]{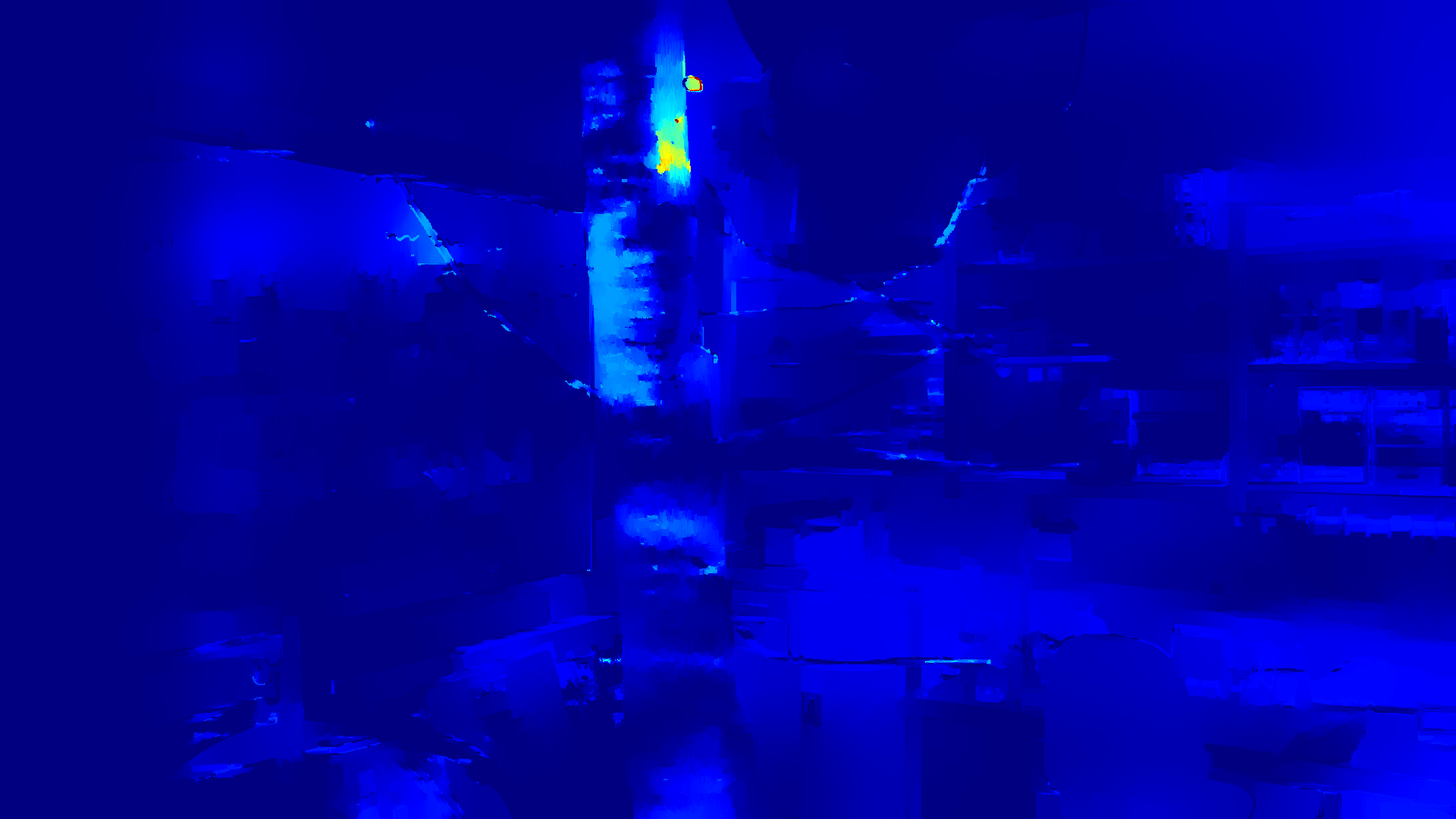}
                \caption{GA optimization with SSIM}
            \end{subfigure}
        \end{minipage}
        }
        \caption{Disparity maps for Example 474 generated using manually
        configured and GA-optimized parameters (transferred from Example 33), demonstrating
        stable generalization capability of GA-optimized parameters.}
        \label{474_example}
    \end{figure}

    We compared disparity maps from GA-optimized versus manually tuned parameters
    (Fig.~\ref{33_example}). While GA parameters achieved stable quantitative metrics,
    manual parameters exhibited highly unstable performance, sometimes appearing
    effective but often failing, highlighting that global optimization metrics
    may compromise fine details in specific regions. To evaluate generalizability,
    we applied both parameter sets from Example 33 to Example 474 under varying
    imaging conditions. As shown in Fig.~\ref{474_example}, manual parameters performed
    poorly while GA-optimized parameters maintained consistent performance.

    This result highlights a critical advantage of our GA-optimization
    approach: while manually tuned parameters may achieve good results for
    specific images, they often fail to generalize across varied imaging conditions.
    In contrast, our GA-optimized parameters demonstrated superior adaptability,
    producing structurally coherent disparity maps across different examples
    with varying image characteristics. Our results demonstrate that GA-based parameter
    optimization effectively improves SGBM+WLS disparity map quality across multiple
    metrics, with superior generalization across varied imaging
    conditions—crucial for dynamic UAV applications.
    \section{Conclusions}
    This study presents a GA-based parameter optimization approach for SGBM+WLS stereo
    matching in UAV applications. Our method systematically explores parameter
    space, achieving 42.86\% MSE reduction, 8.47\% PSNR increase, and 28.52\%
    SSIM improvement compared to baseline configurations. GA-optimized parameters
    demonstrate superior adaptability across varied imaging conditions compared to
    manual tuning. While processing at 0.5 seconds per frame (faster than neural
    networks but not real-time), optimization opportunities exist through C++ implementation,
    GPU acceleration, or parallel processing. For UAV pruning tasks, SSIM-optimized
    parameters provide superior structural representation of branches critical for
    successful operations.
    \section{ACKNOWLEDGEMENT}
    This work was supported in part by the MBIE Endeavor Research
    Programme UOCX2104, the MBIE Data Science SSIF Fund under contract RTVU1914,
    and Marsden Fund of New Zealand Government under Contracts VUW2115.
    \bibliographystyle{IEEEtran}
    \bibliography{references}
\end{document}